%% file: eir.tex
\documentclass[preprint]{article}

    \PassOptionsToPackage{square,  sort, numbers, compress}{natbib}

\usepackage{neurips_2023}

\usepackage[utf8]{inputenc} 
\usepackage[T1]{fontenc}    
\usepackage{hyperref}       
\usepackage{url}            
\usepackage{booktabs}       
\usepackage{amsfonts}       
\usepackage{nicefrac}       
\usepackage{microtype}      
\usepackage{xcolor}         

\usepackage{amsmath}
\usepackage{amssymb}
\usepackage{placeins}
\usepackage{graphicx}
\usepackage{epstopdf}
\usepackage{multirow}
\usepackage{multicol}
\usepackage{subcaption}
\usepackage[nolist]{acronym}
\usepackage[capitalize]{cleveref}
\crefname{section}{Sec.}{Secs.}
\Crefname{section}{Section}{Sections}
\Crefname{table}{Table}{Tables}
\crefname{table}{Tab.}{Tabs.}
\usepackage{xspace}
\makeatletter
\DeclareRobustCommand\onedot{\futurelet\@let@token\@onedot}
\def\@onedot{\ifx\@let@token.\else.\null\fi\xspace}

\def\eg{\emph{e.g}\onedot} 
\def\ie{\emph{i.e}\onedot}

 \def\vs{\emph{vs}\onedot}

\makeatother

\title{Contextualising Implicit Representations for Semantic Tasks}

\author{%
  Theo W. Costain\\
  Active Vision Lab\\
  University of Oxford\\
  \texttt{costain@robots.ox.ac.uk} \\
  \And
  Kejie Li\\
  Active Vision Lab\\
  University of Oxford\\
  \texttt{kejie@robots.ox.ac.uk} \\
  \And
   Victor A. Prisacariu \\
  Active Vision Lab\\
  University of Oxford\\
  \texttt{victor@robots.ox.ac.uk} \\
}

\begin{document}

\begin{acronym}
    \acro{IR}{implicit representation}
    \acro{UDF}{unsigned distance function}
\end{acronym}

\maketitle

\begin{abstract}
    Prior works have demonstrated that implicit representations trained only for reconstruction tasks typically generate encodings that are not useful for semantic tasks.
    In this work, we propose a method that contextualises the encodings of implicit representations, enabling their use in downstream tasks (e.g. semantic segmentation), without requiring access to the original training data or encoding network.
    Using an implicit representation trained for a reconstruction task alone, our contextualising module takes an encoding trained for reconstruction only and reveals meaningful semantic information that is hidden in the encodings, without compromising the reconstruction performance.
    With our proposed module, it becomes possible to pre-train implicit representations on larger datasets, improving their reconstruction performance compared to training on only a smaller labelled dataset, whilst maintaining their segmentation performance on the labelled dataset.
    Importantly, our method allows for future foundation implicit representation models to be fine-tuned on unseen tasks, regardless of encoder or dataset availability.
\end{abstract}

\section{Introduction}
\label{sec:intro}
\input{sections/1_intro}

\section{Related Work}
\label{sec:rel}
\input{sections/2_related}

\section{Method}
\label{sec:meth}
\input{sections/3_method}

\section{Experiments}
\label{sec:expr}
\input{sections/4_experiments}

\section{Results}
\label{sec:result}
\input{sections/5_results}

\section{Limitations}
\label{sec:lim}
\input{sections/6_discussion}

\section{Conclusion}
\label{sec:conc}
\input{sections/7_conclusion}

\FloatBarrier
{\small
    \bibliographystyle{unsrtnat}
    \bibliography{bib}
}

\end{document}

%% file: sections/1_intro.tex
The explosion of interest in augmented reality in recent years has spurred a renewed search for more efficient representations of 3D data.
Whilst point-clouds, meshes, and various other representations have been proposed over the years, the recent introduction of implicit representations like NeRF and DeepSDF have reignited interest in the \emph{representation} rather than the processing of the data.

Opposed to ``classical'' representations that discretise the underlying structure, \acp{IR} learn a continuous function over 3D space.
\acp{IR} are able to represent the structure at arbitrary resolutions, trading spatial complexity for time complexity required to extract the structure from the representation.
In the most current approaches, an encoder takes input in one or more modalities producing an encoding that is used to condition an MLP that composes the function.
Early works, such as DeepSDF\cite{park2019deepsdf} and Occupancy Networks\cite{mescheder2019occupancy}, learned functions that separated space into ``inside'' and ``outside'' regions, however, this ensured that the network could only learn closed surfaces.
Subsequently, methods were proposed that resolved this limitation by learning an \ac{UDF}, where the surface of the object lies on the zero level set of the function.
More work followed, and improved on various aspects of these approaches including training ambiguities~\cite{zhou2022learning,yang2023neural} and extraction/rendering~\cite{zhou2022learning} (further discussion in \cref{sec:rel}), but despite this, relatively little attention was given to applications of these approaches in conventional pipelines.

Foundation models, \ie large pre-trained generalist networks and models (\eg \cite{resnet,openai2023gpt4}), that are trained on vast amounts of data, allowing adaptation to a variety of downstream tasks, are increasingly an essential building block in deep learning pipelines.
It is not impossible that either for safety or commercial reasons (as is already beginning to happen\cite{openai2023gpt4}), foundation like \ac{IR} encoders may not be publicly released, nor their training data, instead only encodings that represent a given shape and the requisite decoder may be made publicly available.
Given \citet{costain2021towards} demonstrated that, when trained for reconstruction tasks only, \acp{IR} learn encodings that are not necessarily meaningful for semantic tasks.
Accordingly, without the ability to train (or re-train) the encoder with semantic supervision\cite{wang2022rangeudf}, it has been observed~\cite{costain2021towards} that the performance on semantic tasks is extremely poor.

To address this problem, we propose a novel method to contextualise the encodings learnt by networks supervised on reconstruction tasks alone, even when the original training data is not available.
Basing our experiments on the approach of \citet{wang2022rangeudf}, we show the limitations of encodings generated by training on reconstruction tasks alone.
Then we propose our lightweight contextualising module that takes the learnt encoding and produces a small additional context encoding.
This context encoding can then be combined with the existing encoding allowing the network to completely recover performance on the semantic tasks.

By separating the geometric tasks from the semantic tasks, our approach allows the geometric pipeline to be trained on much cheaper to produce datasets where complete semantic labels are not available, before our contextualising module is applied to a smaller fully labelled dataset enabling semantic segmentation alongside reconstruction.
Rather than complex approaches~\cite{li2017forgetting}, our method presents a simple but effective and performant approach that address a major shortfall in existing implicit representation approaches.

Our key contributions are:
\begin{itemize}
    \item Our contextualising module which reveals hidden semantic information contained in the feature encodings of \ac{IR}.
    \item A novel and simple approach to train existing implicit representations for unseen semantic tasks without access to the original training data.
\end{itemize}

In the rest of this paper we cover: relevant existing works in the literature \Cref{sec:rel}, our method and contextualising module \Cref{sec:meth}, details of our experimental setup \Cref{sec:expr}, the results of our experiments \Cref{sec:result}, and finally the limitations of our approach \Cref{sec:lim}.

%% file: sections/2_related.tex
Early \ac{IR} works~\cite{mescheder2019occupancy,park2019deepsdf,atzmon2020sal,gropp2020implicit,chen2019learning,michalkiewicz2019deep,poursaeed2020coupling} focused mainly on reconstructing single objects.
Both Occupancy Networks~\cite{mescheder2019occupancy} and IM-Net~\cite{chen2019learning}, learn a function mapping from points in space to the probability that point lies within the object to be reconstructed.
Occupancy Networks further proposed a heirarchical, octree based, extraction method to efficiently extract the mesh.
In contrast, DeepSDF~\cite{park2019deepsdf} learns a function mapping from space to a signed distance function.
Although they proposed an encoder-decoder structure, they also introduced an auto-decoder structure, where the representation encoding is found by freezing decoder and optimising the encoding/embedding.
Scene Representation Networks (SRN)~\cite{sitzmann2019scene} proposed a ``Neural Renderer'' module, which maps from 3D world coordinates to a feature representation of the scene at that location.
Sign Agnostic Learning (SAL)~\cite{atzmon2020sal} proposed to remove the need for signed ground truth information, whilst still learning a signed distance function.
Crucial to this effort is an initialisation scheme that the initial level set was approximately a sphere of some chosen radius.
\citet{gropp2020implicit} introduce the Eikonal Loss term amongst other improvements to the loss function from SAL~\cite{atzmon2020sal}.
These new terms encourage the representation to develop a unit norm gradient, like a metric SDF, and acts as a geometric regularisation over the learned function, improving smoothness and accuracy of the reconstructions.
Later methods~\cite{luigi2023deep} include approaches to better allow networks to represent high frequency information~\cite{tancik2020fourier,sitzmann2020siren}, the former of which is vital to the performance of NeRFs~\cite{mildenhall2020nerf,martin2021nerfw,chan2021pi}.

These early works focused on single object reconstruction, with typically a single encoding or embedding per object.
This limits the scale of objects these representations could represent, a concern later works proposed several solutions to.
Many works arrived at a similar solution to this problem, using either planes~\cite{peng2020convolutional} or grids~\cite{peng2020convolutional,chibane2020implicit,jiang2020local,chabra2020deep}, to improve both the scale and detail of the reconstructions.
Convolutional Occupancy Networks~\cite{peng2020convolutional} make use of planes or grids of features, and IF-Net~\cite{chibane2020implicit} learns learns a hierarchy of multi-scale features, both interpolating between these features at queried locations to predict occupancy probabilities and signed distance function respectively.
Rather than interpolating features, Deep Local Shapes~\cite{chabra2020deep} and \citet{jiang2020local}, learn a grid of encodings, dividing scenes into small simple geometric shapes.

All the above methods share a common trait in separating space into inside \vs outside, however in the case where watertight meshes are not available (as is the case for common 3D Datasets~\cite{scannet,s3dis,scenenn}) training is not possible without complicated pre-processing, or learning overly thick walls.
\citet{chibane2020neural} addressed this issue by learning an \ac{UDF} as well as proposing a gradient based rendering scheme to extract the surface, a requirement given Marching Cubes~\cite{marchingcubes} cannot be applied to \acp{UDF}.
Various works followed in this vein~\cite{ye2022gifs,avidan2022meshudf,tang2021sa,wang2022rangeudf,zhou2022learning,yang2023neural}.
Notably, \citet{avidan2022meshudf,zhou2022learning} who independently proposed an approach to significantly improve the extraction/rendering of \acp{UDF}, by modifying Marching Cubes to look for diverging gradients rather than zero crossings allowing its use on \acp{UDF}.

A number of works have considered semantic tasks alongside NeRFs~\cite{zhi2021inplace,vora2022nesf,yang2021learning,wu2022object,kundu2022panoptic}, however far fewer works\cite{kohli2020semantic,costain2021towards,luigi2023deep,wang2022rangeudf} consider semantic tasks alongside \acp{IR}.
\citet{costain2021towards} argued that training \acp{IR} on geometric tasks alone produce encodings that are poor for semantic tasks.
However, \citet{luigi2023deep} show that these encodings still contain the semantic information, and that it is possible to transform these encodings into a form that are more meaningful for semantic tasks.
We leverage this insight in designing our contextualising module.
\citet{wang2022rangeudf}, as well as proposing a \ac{UDF} based \ac{IR}, train a ``surface-aware'' segmentation branch alongside the \ac{UDF}.

As a fundamental problem in computer vision, a vast array of works~\cite{he2021deep,lin2020fpconv,qi2017pointnet,qi2017pointnet++,wu2019pointconv,zhao2012pointtransformer,hu2020randla,rethage2018fully,riegler2017octnet,graham20183d,qiu2020dgcn,thomas2019kpconv,hanocka2019meshcnn,xu2017directionally,dai2018scancomplete,li2018pointcnn,hua2018pointwise} have tackled semantic segmentation of point clouds, however a detailed discussion of these methods falls outside the scope of this work.

%% file: sections/3_method.tex
\input{figures/overview.tex}

Implicit representations seek to learn a functional mapping, $f$, from a query point, $q\in\mathbb{R}^3$, in space to the distance from that query point to the nearest point on the surface being represented.
In this work we consider \acp{UDF}, further constraining $f:q\in\mathbb{R}^3 \mapsto \mathbb{R}_0^+$.
In this work, we use UDFs, as these are the currently preferred way to represent non watertight scenes, however, this should not affect the generality of the approach, and should a dataset containing large watertight scenes be released, we expect our method should also apply.
This function is typically implemented as a simple MLP, but to avoid overfitting the MLP to every surface to be represented, it is often desirable to condition the function on some global~\cite{mescheder2019occupancy,park2019deepsdf} or local~\cite{wang2022rangeudf,chibane2020neural} encoding of shape, giving $f:q\in\mathbb{R}^3, E\in\mathbb{R}^d \mapsto \mathbb{R}_0^+$, where $E$ is some encoding vector and $d$ its dimension.

In our work, as the only method that performs semantic tasks alongside learning implicit representations for large scenes, we use the RangeUDF method proposed by \citet{wang2022rangeudf}.
Their approach takes an input point-cloud $P\in\mathbb{R}^{N\times3}$, where $N$ is the number of points, and uses an encoder to learn some encoded features, $E_g\in\mathbb{R}^{N\times d}$.
These encoded features are then passed to the decoder(s), alongside a set of query points $Q\in\mathbb{R}^{M\times3}$ (where $M$ is the number of query points), where they use KNN to collect the $K$ nearest encoding vectors and combine them using a simple attention module which is then fed into their \ac{UDF} and semantic segmentation module.

\subsection{The Problem}

A common pipeline in computer vision tasks is to take a pre-trained model, that produces meaningful features for a given task, and either fine-tune it, or use the generated features as input to another module that performs some desired task.
The arc of research so far has resulted in this pre-training often\cite{resnet} taking the form of classification tasks on extremely large datasets~\cite{imagenet}, ensuring these pre-trained models learn features that are semantically meaningful.

On the other hand, \ac{IR} methods have arisen to tackle a different challenge: the representation of 3D shape and structure.
Typically, this is in service of reducing the memory required to represent a given scene or object at high resolutions~\cite{peng2020convolutional,mescheder2019occupancy,park2019deepsdf}, compared to other conventional representations such as point-clouds, meshes, or voxel grids.
However, much of the research on implicit representations to date has focused on the reconstruction task alone, with little consideration of how they might be used to replace conventional representations in existing pipelines.

When training the encodings that condition the \ac{UDF}, the desire is for the network to learn some set of encodings $E$ that holistically represent local structural information about the underlying shape.
Our experiments in \cref{sec:results:tgnir}, confirm~\cite{costain2021towards} that the encodings, $E_g$, learnt when training the \ac{UDF} alone for geometric reconstruction show poor separability in semantic space (\cref{fig:tsne:a}).
Whilst this can obviously be addressed by training for both semantic and reconstruction tasks jointly~\cite{wang2022rangeudf,costain2021towards}, it is trivial to imagine scenarios where it is extremely desirable to be able to fine-tune on semantic tasks, without requiring either access to the original training data (original training data may not be publicly available) or having to potentially expensively retrain the entire pipeline.
It also bears noting that whilst it is possible to exhaustively render/extract each scene from the encoding and UDF, and use this to re-create the training data, current implicit representation methods are far from perfect, and so taking this approach would almost certainly compound errors, not to mention the substantial cost of labelling the extracted scenes. Our proposed method avoids this entire problem, with a simple process.

\input{figures/tsne.tex}

\subsection{Contextualising Module}

The results of \citet{zhou2022learning} suggest that although not necessarily in present in a separable form, the semantic information is still present in the representation.
Accordingly, we propose our simple contextualising module, which produces \emph{compact} context features that carry substantial semantic information (\cref{fig:tsne:b}), that when combined with the original encoded features, produces features useful for semantic segmentation as well as reconstruction.
An overview of our method is presented in \cref{fig:over}.

Taking the encoded features, our module uses a small encoder-decoder UNet-like network, specifically PointTransformer\cite{zhao2012pointtransformer}, to re-capture semantic information present in the encoding.
Important to its function is the contextualising modules ability to consider wider shape context, than either the \ac{UDF} or segmentation decoder, which has repeatedly been shown as vital to capturing semantic information\cite{qi2017pointnet,qi2017pointnet++}.
This re-capturing of a wider scene context gives rise to the naming of our module.
This is achieved through the PointTransformer's downsampling, interpolation, and upsampling performed across 5 different scales (similar to \cite{qi2017pointnet++}).

The formulation of RangeUDF which predicts the semantic class, $s_i\in\mathbb{R}^C$ where $C$ is the number of classes, of a given point $q_i\in Q$ as
\begin{equation}
    s_i = f_{\text{sem}}(q_i|E_g)
\end{equation}
Instead, our contextualising module, $f_{\text{ctx}}$, takes the fixed encoded features (trained on only the reconstruction task), $E_g\in\mathbb{R}^{M\times d}$, and predicts a set of context features, $E_c\in\mathbb{R}^{M\times l}$.
We then concatenate the context features with the original encoded features to give the semantic features, $E_s\in\mathbb{R}^{M\times(d+l)}$, which we feed into the segmentation module alongside the query points, giving instead
\begin{equation}
    s_i = f_{\text{sem}}(q_i|E_g \oplus f_{\text{ctx}}(E_g)) = f_{\text{sem}}(q_i|E_s)
\end{equation}
where $\oplus$ represents concatenation in the feature dimension.

Despite the simplicity of our contextualising module and its implementation, our results demonstrate the performance improvements it provides to the semantic task are substantial.
Our contextualising module is implemented as a substantially shrunk version of the PointTransformer, reducing the number of parameters from roughly 7.8 million to around 379,000.
This is achieved through a reduction of the number of channels at each scale from $[32, 64, 128, 256, 512]$ to $[32, 32, 64, 64, 128]$ and reducing the number of ``blocks'' at each scale to 1.

During training, we use the L1 loss, with the same clamping as \citet{chibane2020neural}, for supervising the reconstruction task, and the standard cross entropy loss for the segmentation task.
Our method focuses on mainly on separately training each task, in which case, our loss contains only a single objective, avoiding the need to balance loss terms entirely.
However, in the case of joint training, following \cite{wang2022rangeudf} we use the uncertainty loss~\cite{kendall2018multi} to avoid manually tuning loss weightings.

%% file: figures/overview.tex
\begin{figure}[t]
    \centering
    \includegraphics[width=0.95\linewidth]{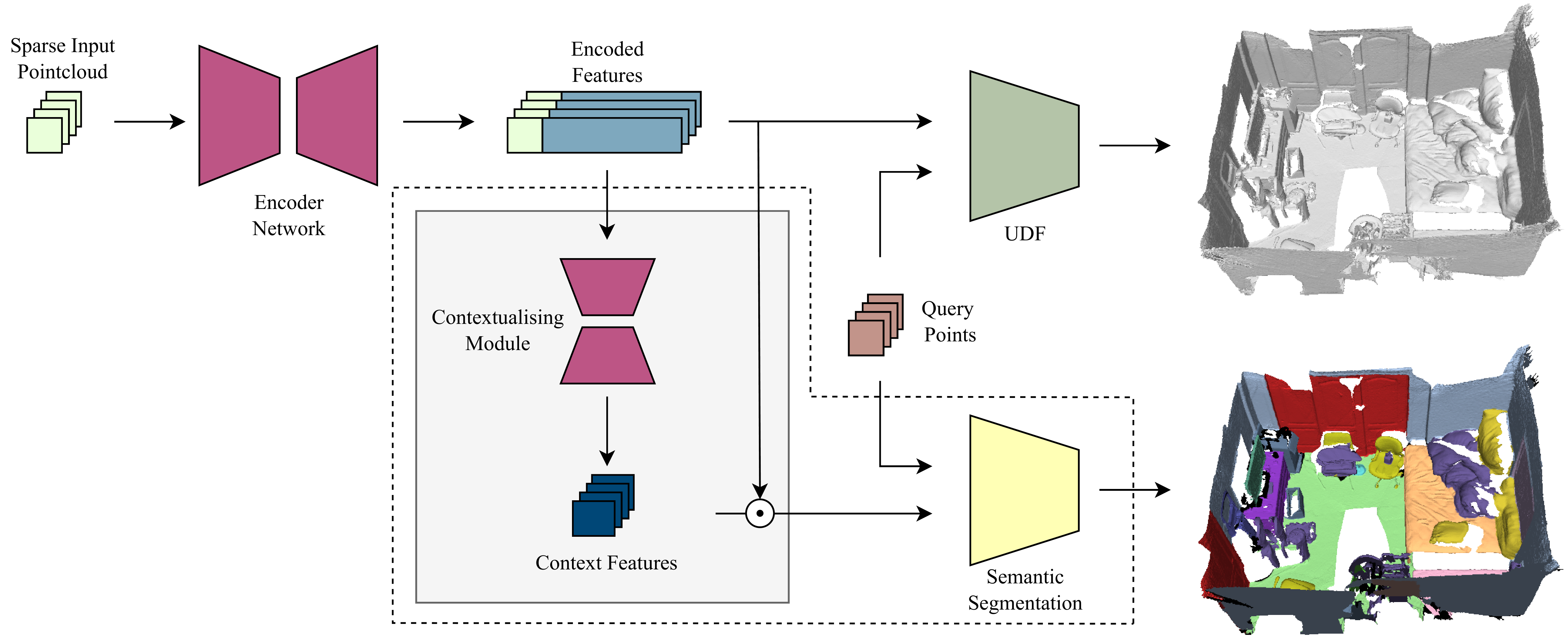}
    \caption{Our proposed contextualising module (light grey box) allows already trained implicit representations, to be fine-tuned (training only elements inside the dotted line) on semantic tasks without the need for the original training data. Learning a compact contextualising vector which is concatenated with the original encoding, our module allows full semantic performance to be recovered from encoders trained only on reconstruction tasks.}
    \label{fig:over}
\end{figure}

%% file: figures/tsne.tex
\begin{figure}[tb]
    \centering
    \def \tsneimagesize {0.24}
    \subcaptionbox{\label{fig:tsne:a}Geometry trained }[\tsneimagesize\linewidth]{\includegraphics[width=0.99\linewidth]{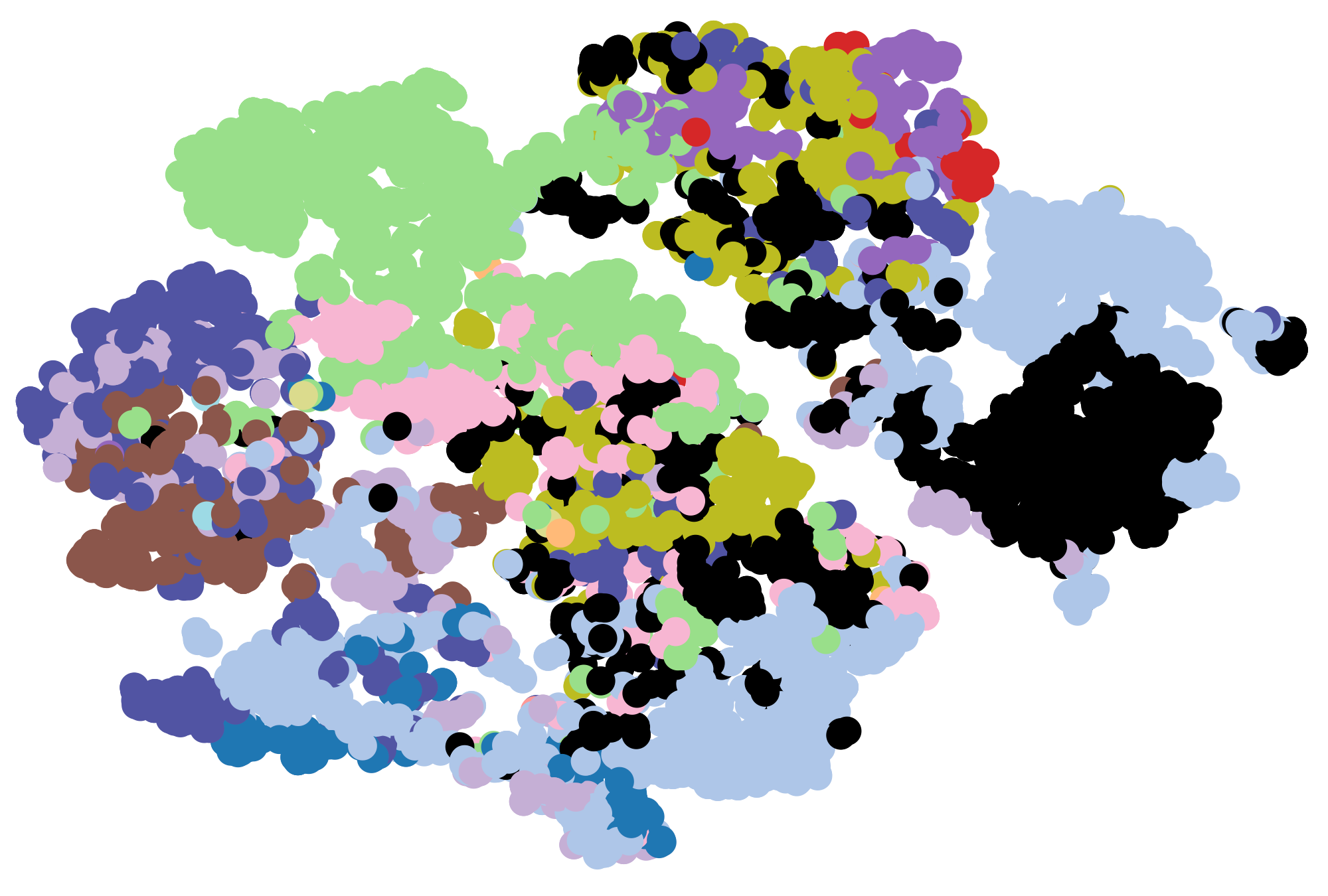}}\hfill%
    \subcaptionbox{\label{fig:tsne:b}Context features}[\tsneimagesize\linewidth]{\includegraphics[width=0.99\linewidth]{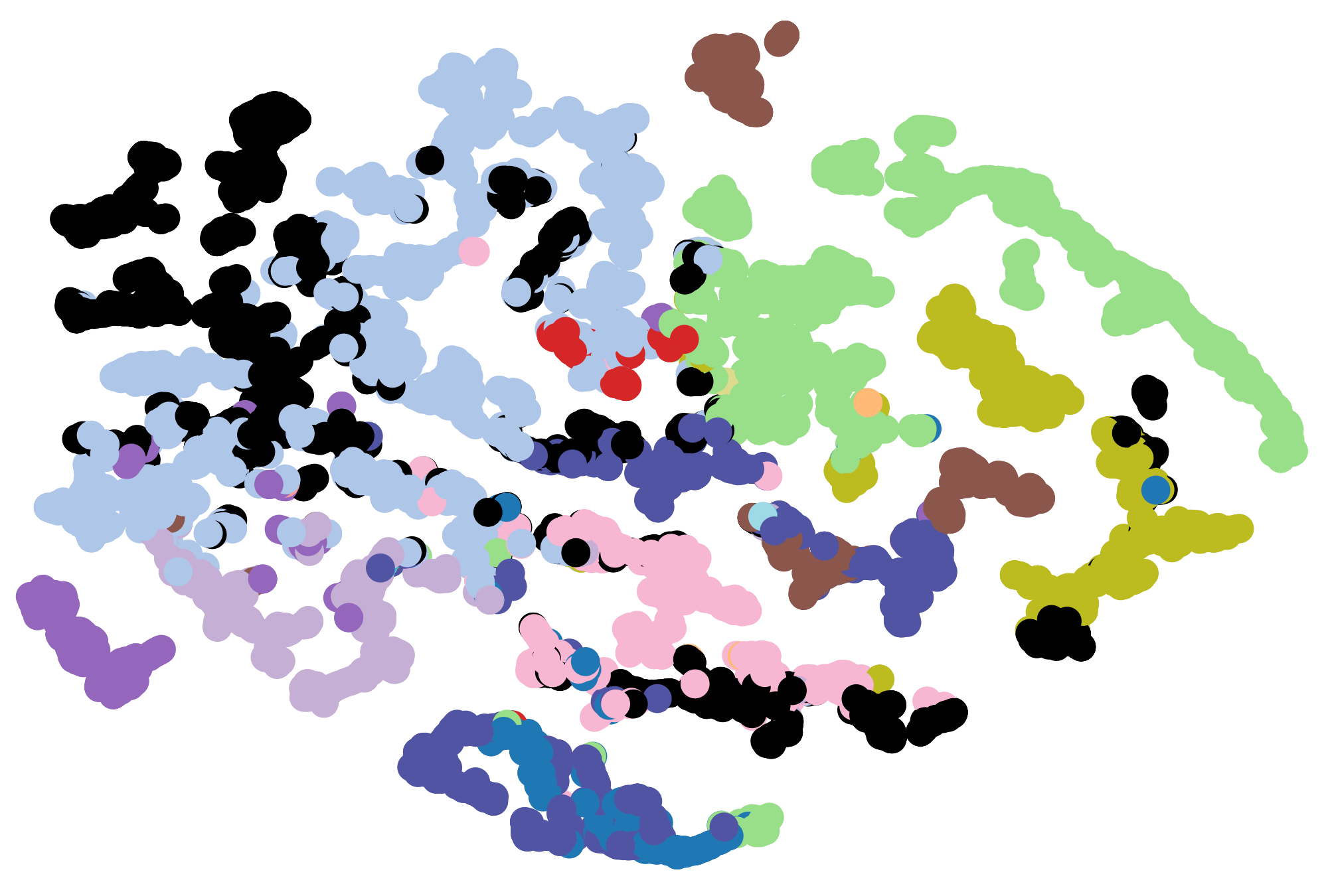}}\hfill%
    \subcaptionbox{\label{fig:tsne:c}\centering Concatenated geometry and context}[\tsneimagesize\linewidth]{\includegraphics[width=0.99\linewidth]{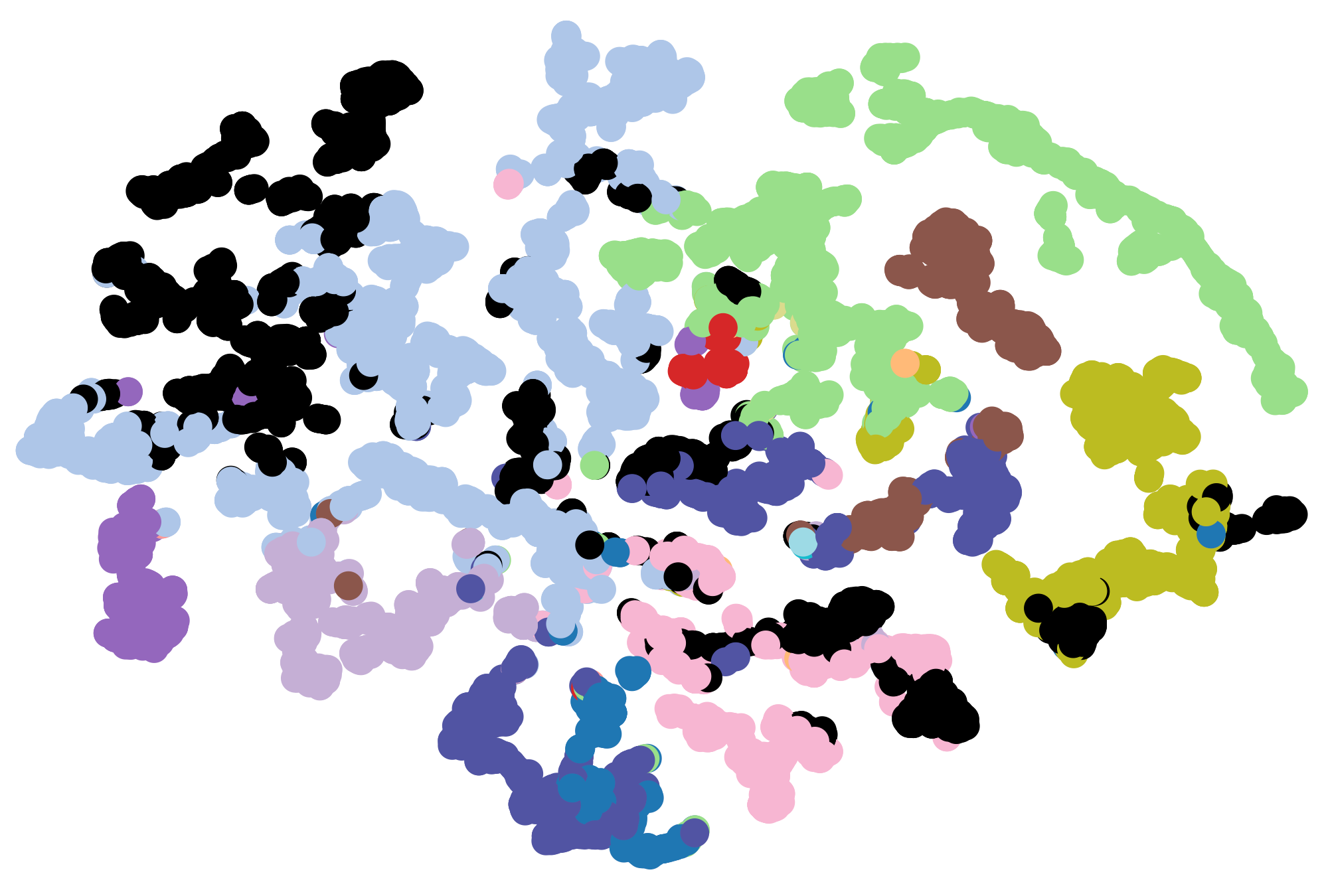}}\hfill%
    \subcaptionbox{\label{fig:tsne:d}\centering Joint semantic \& reconstruction trained}[\tsneimagesize\linewidth]{\includegraphics[width=0.99\linewidth]{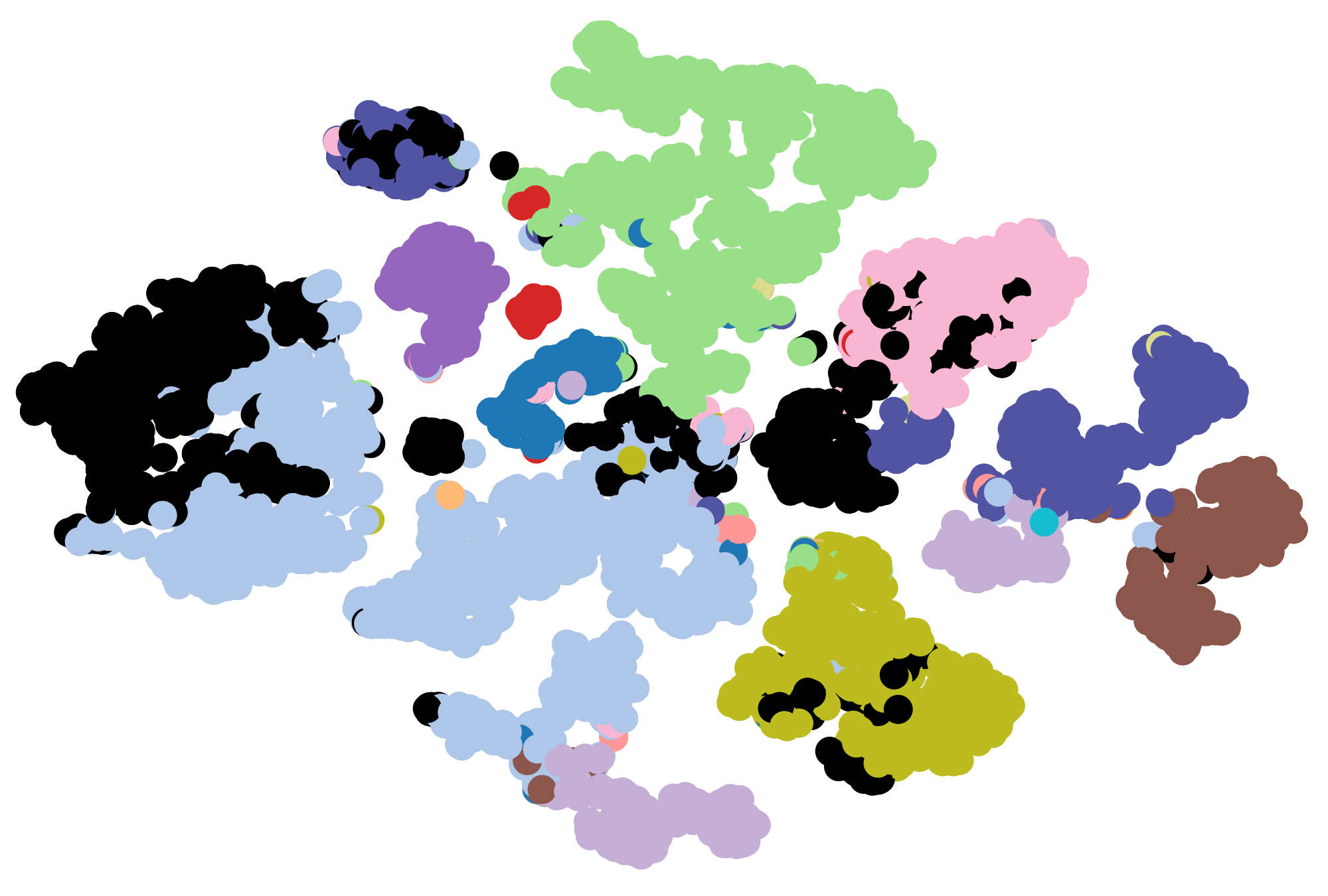}}
    \caption{t-SNE embeddings of the features of a scene encoding.
        The features trained for geomery tasks only, show poor separability according to semantic label.
        Our proposed contextualising module produces context features that are clearly more separable, and become even more so when combined with the existing features. Figure best viewed in colour.
    }
    \label{fig:tsne}
\end{figure}

%% file: sections/4_experiments.tex
In this section, we cover details of the datasets and metrics used in our experiments, as well as the relevant details of our implementations and the resources used to perform our experiments.

\subsection{Datasets \& Metrics}
We train and evaluate our method on three datasets: ScanNet~\cite{scannet}, SceneNN~\cite{scenenn}, and 2D-3D-S~\cite{s3dis}, all of which are captured using RGB-D cameras.

\textbf{ScanNet}
The ScanNet dataset consists of 1613 scans of real-world rooms.
The data is split into 1201 scans for training and 312 scans for validation with a further 100 scans held out for online benchmarks.
Following \citet{wang2022rangeudf}, we use the validation set for testing.
Semantic labels are provided for 40 classes, however following other methods~\cite{qi2017pointnet,qi2017pointnet++,wang2022rangeudf,thomas2019kpconv,wu2019pointconv,hu2020randla,lin2020fpconv}, we train and test on only the 20 class subset used in the online benchmark.

\textbf{2D-3D-S}
The 2D-3D-S dataset consists of 6 \emph{very} large-scale indoor scans, capturing rooms, hallways and other educational and office like environments using an RGB-D
The data is divided into a total of 271 rooms, divided into 6 ``Areas'' based on the scan they are contained in.
Area 5 is split into two scans without a provided registration between them, preventing their use in the data preparation pipeline described below.
Following \citet{wang2022rangeudf}, we use Areas 1-4 for training and Area 6 for testing.
The semantic labels are provided for 13 classes.

\textbf{SceneNN}
The SceneNN dataset consists of 76 indoor scans divided into 56 scenes for training and 20 scenes for testing~\cite{hua2018pointwise}.
Semantic labels are provided for the same 40 classes as ScanNet, where again we use the 20 class subset.

\subsubsection{Data Preparation}
We follow the same processing steps as \citet{chibane2020neural}, normalising each scene's mesh to a unit cube, and sampling 10k surface points (for the encoder input) and 100k off surface points for which we compute the distance to the closest point on the surface.

\subsubsection{Metrics}
When evaluating the reconstruction tasks, we use the standard Chamfer L1 \& L2 distance measures (lower is better) as well as the F1-$\delta$ and F1-$2\delta$ score (higher is better).
All CD-L1 values are reported $\times10^{-2}$ and CD-L2 values $\times10^{-4}$, and we set $\delta=0.005$.
For the segmentation task, we use mean Intersection-over-Union (higher is better) as well as mean F1-$\delta$, which is calculated by determining the per-class F1-$\delta$ score then averaging over the classes.

\subsection{Implementation Details}
We implement our work in PyTorch~\cite{pytorch}, and perform our experiments on 3 Nvidia RTX6000 GPUs and an Intel Xenon Gold 6226R CPU.
We use the Adam optimiser with default parameters and a learning rate of $10^{-3}$ for all experiments, we use a batch size of 12, and set the dimension of the context features to 4.
During training, we feed 10,240 points into the encoder, and 50k points to the \ac{UDF} and segmentation decoder.
For experiments on ScanNet, we train the model for 500 epochs. For both 2D-3D-S and SceneNN, we train for 1k epochs.

For the encoder network, we use the PointTransformer~\cite{zhao2012pointtransformer} network.
To drastically speed up the evaluation of our experiments, we use the surface extraction algorithm from \citet{zhou2022learning} rather than Algorithm 1 from \citet{chibane2020neural}.

%% file: sections/5_results.tex
\paragraph{Comparing reconstruction-only trained features with our contextualised features}
\label{sec:results:tgnir}
\input{figures/qual.tex}

We start with our experiments confirming that the findings of \cite{costain2021towards} apply to larger implicit representations.
To test this, we train only the encoder network and UDF on the reconstruction task for a given dataset (2\textsuperscript{nd} row \cref{tab:scannet_tgnir:a,tab:scannet_tgnir:b,tab:scannet_tgnir:c}).
Then, freezing the encoder network and UDF, we train the segmentation decoder on the semantic labels of the same dataset, using the frozen encodings (3\textsuperscript{rd} row \cref{tab:scannet_tgnir:a,tab:scannet_tgnir:b,tab:scannet_tgnir:c}).

Qualitative results are shown in \cref{fig:qual}, where the middle left shows the baseline results, and middle right shows the frozen encoder results.
Its clear from the mIOU and mF1 scores that the frozen encodings are insufficient for reasonable quality segmentation results, with the frozen encodings giving roughly half the performance of the baseline jointly trained model.

\input{tables/scannet.tex}

Next, we train the segmentation decoder with the frozen encodings combined with the context features produced by our contextualising module (4\textsuperscript{th} row \cref{tab:scannet_tgnir:a,tab:scannet_tgnir:b,tab:scannet_tgnir:c}).
Again, the mIOU and mF1 scores show that our contextualising module allows nearly full performance on the segmentation task to be recovered, and surprisingly, substantially improves performance over the joint training baseline in the case of ScanNet.
We suspect this improvement arises from the geometric-only encodings' superior reconstruction performance compared to the baseline, as better representation of fine-detailed structure in turn allows better segmentation of these same fine-detailed structures.
In turn, we also suspect the improvement in the geometric-only encodings' results might arise from the reconstruction task being relatively simpler to learn than the segmentation task.

\paragraph{Cross Training \& Validation}
\input{tables/ctx_scannet.tex}
To evaluate one of the key advantages of our proposed method, we cross-train fixed reconstruction-only trained feature encodings with our contextualising module on each of the datasets.
We also train and additional set of fixed encodings on the amalgamation of the three datasets, which we refer to as the Triad dataset.
To preserve train-test splits, the train split for Triad is sum of the three training splits, and likewise for the validation splits.

Our results in \cref{tab:scannet_ctx} demonstrate that our method not only allows cross training between different datasets for reconstruction and semantic tasks, but \emph{importantly}, our results in the 2\textsuperscript{nd} and 3\textsuperscript{rd} rows in \cref{tab:scannet_ctx:d} show that by leveraging the ability to train for reconstruction on larger datasets and then then semantics on a different smaller dataset, we can maintain the same semantic performance, as a jointly trained baseline, whilst improving the quality of the reconstructions generated.

\subsection{Ablations}
\input{tables/ablations.tex}
To validate our design of the contextualising module and its compactness, we perform ablation experiments on the structure of both our contextualising module, as well as the context features we generate.
For our ablation experiments, we use the 2D-3D-S dataset, all parameters are kept the same as in the above experiments except for the modifications described below.

In our experiments (\cref{tab:abl}), we evaluate the following:
\paragraph{Contextual information}
To confirm that information from across the whole feature encodings for a given shape is vital to our contextualising module, rather than individual feature vectors, we implement our contextualising module as an MLP first, and second as a shallower version of the PointTransformer normally used.
Our results show that the MLP provides little advantage over the raw fixed encodings, and that whilst the shallower PointTransformer recovers some of the performance, there is still a gap in performance compared to the baseline.
These results demonstrate the importance of capturing context across the whole encoding in our proposed module.
\paragraph{Compactness}
To show that our contextualising module is as compact as possible whilst maintaining performance, we evaluate the effects of increasing the size of the contextualising module, either by increasing the number of blocks at each scale, or by increasing the number of channels at each scale, or both simultaneously.
We also demonstrate that further reducing the dimension of the context features below 4 harms the performance of the contextualising module.
However, this specific parametrisation applies only to the datasets we use, and may be different for more complex or simpler datasets.

\label{sec:results:abl}

%% file: figures/qual.tex
\begin{figure}[tb]
    \centering
    \includegraphics[width=.98\linewidth]{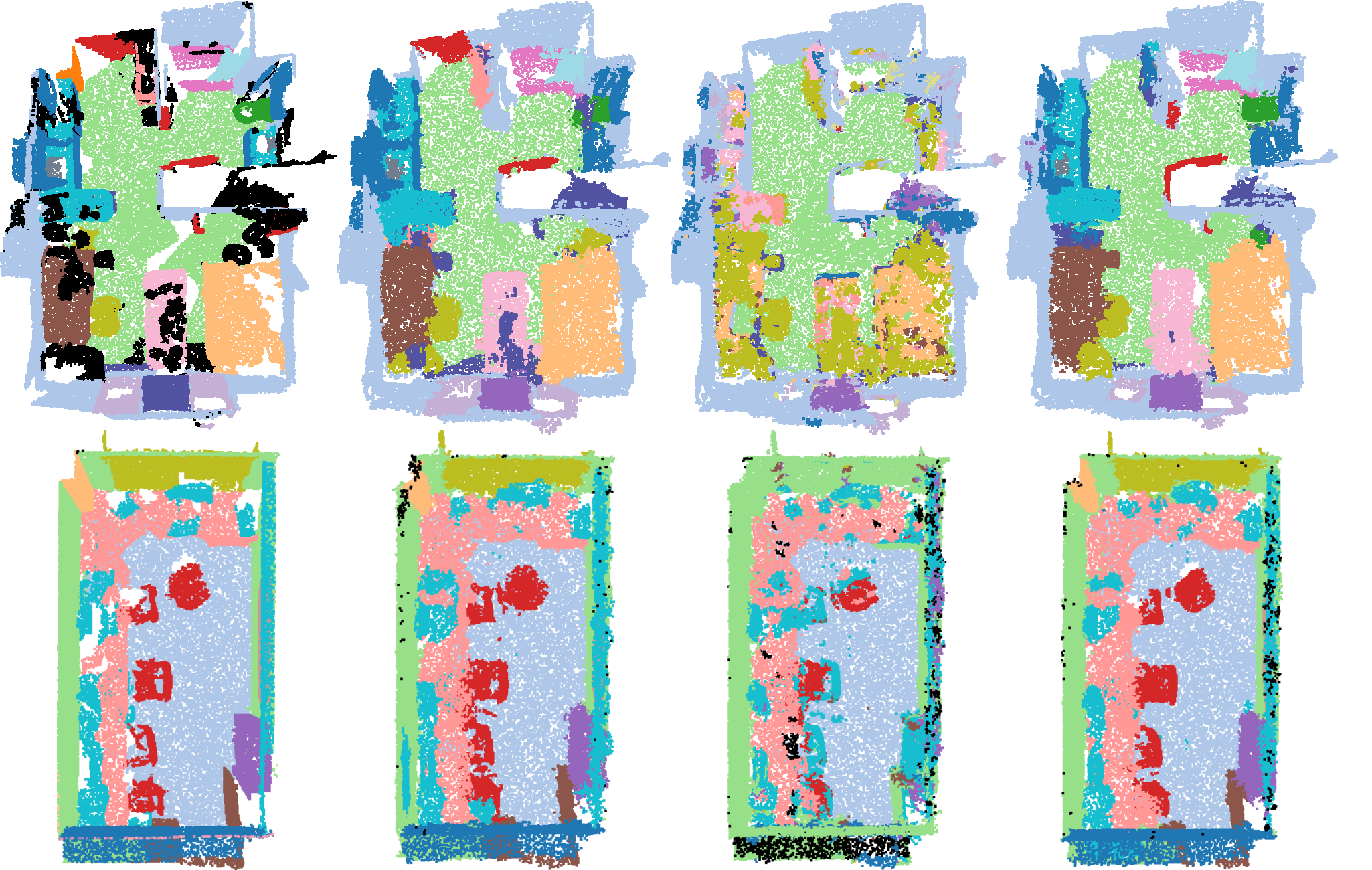}%

    \caption{Qualitative comparison of segmentation and reconstruction on the ScanNet dataset. From left to right: Ground truth (semantics and geometry), jointly trained geometry and segmentation, segmentation training on frozen encodings, and finally segmentation training on frozen encodings using our contextualising module. The frozen encodings trained only for reconstruction seriously inhibit the network's performance on segmentation.
    }
    \label{fig:qual}
\end{figure}

%% file: tables/scannet.tex
\begin{table}[tb]
    \centering
    \setlength{\tabcolsep}{4pt}
    \subcaptionbox{\label{tab:scannet_tgnir:a}ScanNet}{
        \begin{tabular}{l@{\hskip 1em}ccccccc}\toprule
                             & L1 $(\downarrow)$ & L2 $(\downarrow)$ & F1-$\delta$ $(\uparrow)$ & F1-$2\delta$ $(\uparrow)$ & mF1-$\delta$ $(\uparrow)$ & mF1-$2\delta$ $(\uparrow)$ & mIOU $(\uparrow)$ \\
            \midrule
            Baseline         & 0.312             & 0.183             & 0.872                    & 0.962                     & 0.487                     & 0.522                      & 0.536             \\
            Geometric Only   & 0.302             & 0.176             & 0.884                    & 0.963                     & -                         & -                          & -                 \\
            Frozen Encoder   & 0.298             & 0.170             & 0.888                    & 0.964                     & 0.280                     & 0.3052                     & 0.296             \\
            Context Features & 0.297             & 0.170             & 0.889                    & 0.964                     & 0.640                     & 0.668                      & 0.694             \\

            \bottomrule
        \end{tabular}
    }
    \subcaptionbox{\label{tab:scannet_tgnir:b}SceneNN}{
        \begin{tabular}{l@{\hskip 1em}ccccccc}\toprule
                             & L1 $(\downarrow)$ & L2 $(\downarrow)$ & F1-$\delta$ $(\uparrow)$ & F1-$2\delta$ $(\uparrow)$ & mF1-$\delta$ $(\uparrow)$ & mF1-$2\delta$ $(\uparrow)$ & mIOU $(\uparrow)$ \\
            \midrule
            Baseline         & 0.859             & 5.62              & 0.603                    & 0.836                     & 0.604                     & 0.655                      & 0.692             \\
            Geometric Only   & 0.832             & 5.70              & 0.633                    & 0.849                     & -                         & -                          & -                 \\
            Frozen Encoder   & 0.864             & 6.22              & 0.624                    & 0.844                     & 0.575                     & 0.603                      & 0.600             \\
            Context Features & 0.793             & 4.80              & 0.631                    & 0.849                     & 0.562                     & 0.610                      & 0.648             \\
            \bottomrule
        \end{tabular}
    }
    \subcaptionbox{\label{tab:scannet_tgnir:c}2D-3D-S}{
        \begin{tabular}{c@{\hskip 1em}ccccccc}\toprule
                             & L1 $(\downarrow)$ & L2 $(\downarrow)$ & F1-$\delta$ $(\uparrow)$ & F1-$2\delta$ $(\uparrow)$ & mF1-$\delta$ $(\uparrow)$ & mF1-$2\delta$ $(\uparrow)$ & mIOU $(\uparrow)$ \\
            \midrule
            Baseline         & 0.364             & 0.237             & 0.819                    & 0.960                     & 0.695                     & 0.809                      & 0.727             \\
            Geometric Only   & 0.389             & 0.420             & 0.822                    & 0.951                     & -                         & -                          & -                 \\
            Frozen Encoder   & 0.358             & 0.255             & 0.837                    & 0.960                     & 0.458                     & 0.537                      & 0.435             \\
            Context Features & 0.357             & 0.252             & 0.838                    & 0.960                     & 0.684                     & 0.784                      & 0.700             \\
            \bottomrule
        \end{tabular}
    }
    \caption{Comparison of semantic segmentation and geometric reconstruction, on three datasets. The rows from top to bottom: joint training baseline, geometry reconstruction supervision only, semantic training on frozen encodings from geometry only, semantic training on froxen encodings with our contextualising module.
    }
    \label{tab:scannet_tgnir}
\end{table}

%% file: tables/ctx_scannet.tex
\begin{table}[tb]
    \centering
    \setlength{\tabcolsep}{4pt}
    \subcaptionbox{ScanNet trained geometric features\label{tab:scannet_ctx:a}}{
        \begin{tabular}{lccccccc}\toprule
                            & L1 $(\downarrow)$ & L2 $(\downarrow)$ & F1-$\delta$ $(\uparrow)$ & F1-$2\delta$ $(\uparrow)$ & mF1-$\delta$ $(\uparrow)$ & mF1-$2\delta$ $(\uparrow)$ & mIOU $(\uparrow)$ \\
            \midrule
            ScanNet Labels  & 0.297             & 0.170             & 0.889                    & 0.964                     & 0.639                     & 0.668                      & 0.694             \\
            SceneNN Labels  & 0.344             & 0.221             & 0.836                    & 0.954                     & 0.584                     & 0.617                      & 0.621             \\
            Stanford Labels & 0.328             & 0.177             & 0.861                    & 0.972                     & 0.692                     & 0.782                      & 0.693             \\

            \bottomrule
        \end{tabular}
    }
    \subcaptionbox{SceneNN trained geometric features\label{tab:scannet_ctx:b}}{
        \begin{tabular}{lccccccc}\toprule
                            & L1 $(\downarrow)$ & L2 $(\downarrow)$ & F1-$\delta$ $(\uparrow)$ & F1-$2\delta$ $(\uparrow)$ & mF1-$\delta$ $(\uparrow)$ & mF1-$2\delta$ $(\uparrow)$ & mIOU $(\uparrow)$ \\
            \midrule
            ScanNet Labels  & 0.781             & 4.94              & 0.681                    & 0.857                     & 0.563                     & 0.611                      & 0.65              \\
            SceneNN Labels  & 0.793             & 4.80              & 0.631                    & 0.849                     & 0.562                     & 0.610                      & 0.648             \\
            Stanford Labels & 0.740             & 3.03              & 0.626                    & 0.875                     & 0.5578                    & 0.719                      & 0.679             \\

            \bottomrule
        \end{tabular}
    }
    \subcaptionbox{2D-3D-S trained geometric features\label{tab:scannet_ctx:c}}{
        \begin{tabular}{lccccccc}\toprule
                            & L1 $(\downarrow)$ & L2 $(\downarrow)$ & F1-$\delta$ $(\uparrow)$ & F1-$2\delta$ $(\uparrow)$ & mF1-$\delta$ $(\uparrow)$ & mF1-$2\delta$ $(\uparrow)$ & mIOU $(\uparrow)$ \\
            \midrule
            ScanNet Labels  & 0.357             & 0.519             & 0.852                    & 0.944                     & 0.623                     & 0.656                      & 0.690             \\
            SceneNN Labels  & 0.379             & 0.300             & 0.806                    & 0.940                     & 0.611                     & 0.645                      & 0.648             \\
            Stanford Labels & 0.357             & 0.252             & 0.838                    & 0.960                     & 0.684                     & 0.784                      & 0.700             \\

            \bottomrule
        \end{tabular}
    }
    \subcaptionbox{Triad trained geometric features\label{tab:scannet_ctx:d}}{
        \begin{tabular}{lccccccc}\toprule
                            & L1 $(\downarrow)$ & L2 $(\downarrow)$ & F1-$\delta$ $(\uparrow)$ & F1-$2\delta$ $(\uparrow)$ & mF1-$\delta$ $(\uparrow)$ & mF1-$2\delta$ $(\uparrow)$ & mIOU $(\uparrow)$ \\
            \midrule
            ScanNet Labels  & 0.299             & 0.174             & 0.887                    & 0.964                     & 0.634                     & 0.664                      & 0.690             \\
            SceneNN Labels  & 0.343             & 0.224             & 0.837                    & 0.954                     & 0.593                     & 0.626                      & 0.631             \\
            Stanford Labels & 0.325             & 0.177             & 0.863                    & 0.971                     & 0.728                     & 0.820                      & 0.729             \\

            \bottomrule
        \end{tabular}
    }

    \caption{Cross training and validation using our contextualising module.
        For each table, we use the fixed feature encodings trained on reconstruction only for one dataset, and then train for segmenation with our contextualising module on each of the datasets.
        Triad represents the amalgamation of all three of the datasets.
    }
    \label{tab:scannet_ctx}
\end{table}

%% file: tables/ablations.tex
\begin{table}[tb]
    \centering

    \begin{tabular}{lcc}
        \toprule
                                                  & F1-$\delta$ $(\uparrow)$ & F1-$2\delta$ $(\uparrow)$ \\
        \midrule
        MLP Conxtext Module                       & 0.490                    & 0.568                     \\
        Shallower network (3 scales)              & 0.644                    & 0.745                     \\[0.5em]
        More blocks ($[1, 2, 3, 5, 2]$)           & 0.669                    & 0.768                     \\
        More channels ($[32, 64, 128, 256, 512]$) & 0.662                    & 0.763                     \\
        More blocks \& channels                   & 0.664                    & 0.765                     \\[0.5em]
        $l=2$                                     & 0.606                    & 0.706                     \\
        $l=1$                                     & 0.566                    & 0.662                     \\[0.5em]
        Ours                                      & 0.664                    & 0.763                     \\

        \bottomrule
    \end{tabular}
    \caption{Results of our ablation experiments on the 2D-3D-S dataset.}
    \label{tab:abl}
\end{table}

%% file: sections/6_discussion.tex
Although Initial convergence of semantic segmentation performance when training the context module is faster than the baseline joint training,  90\% of the performance with 17\% of the training time, full convergence is not materially faster than the joint training.
We suspect, however, this slowness arises from the segmentation module proposed in \cite{wang2022rangeudf}, as training the encoder used to generate the encoded features on a simple segmentation task converges substantially faster.

Ultimately, the main limitation of our approach is that it requires labelled data to train the semantic branch.
However, as our approach separates the training of the reconstruction and semantic tasks, it is theoretically possible to extract meshes from the decoders at a coarse scale, and then manually label them to train the network for semantic tasks.
There are also a number of weaknesses that arise from in baseline RangeUDF~\cite{wang2022rangeudf}, however these improvements would not necessarily represent any novelty, rather incremental improvements that would improve numerical performance, such as replacing their scalar attention module with vector attention or adding positional encoding~\cite{tancik2020fourier} top the decoders.

%% file: sections/7_conclusion.tex
In this work, we propose a novel approach to training implicit representations for downstream semantic tasks without needing access to the original training data or encoding network.
We introduce our contextualising module that reveals semantic information contained in the encodings of implicit representations trained only for geometric tasks.
We demonstrate our contextualising module on the task of semantic segmentation and show that without it, the encoded features learnt by implicit representations for geometric tasks lack sufficient separability to provide meaningful results.
Further, we show that using our module, it becomes possible to leverage larger unlabelled datasets to pre-train implicit representations to and then fine-tune on smaller labelled semantic datasets, achieving higher reconstruction performance than would be possible with only the smaller labelled datasets.